%% file: main.tex
\documentclass{article}

\usepackage[square,numbers]{natbib}

\usepackage[final,nonatbib]{styles/corl_2026} 

\makeatletter
\renewcommand{\@noticestring}{%
  The Science of Physical AI Safety (SPAIS) Workshop at the
  \@conferenceordinal\/ Conference on Robot Learning (CoRL \@conferenceyear),
  \@conferencelocation.%
}
\hypersetup{pdfsubject={SPAIS Workshop at the \@conferenceordinal\/ Conference
  on Robot Learning (CoRL \@conferenceyear)}}
\hypersetup{pdftitle={From Concepts to Actual Causes: Toward Closing the
  Concept-Malfunction Gap in Robot Foundation Models}, pdfauthor={Gihyuk Ko}}
\makeatother


\usepackage{amsmath,amssymb}
\usepackage{booktabs}
\usepackage{tabularx}
\usepackage{wrapfig}
\usepackage{enumitem}
\usepackage{microtype}
\usepackage{tikz}
\usepackage{pifont}
\usetikzlibrary{arrows.meta}

\newcommand{\query}{\mathcal{Q}}

\newcommand{\cmark}{\textcolor{green!50!black}{\ding{51}}}
\newcommand{\xmark}{\textcolor{red!75!black}{\ding{55}}}

\title{From Concepts to Actual Causes:\\ Toward Closing the Concept--Malfunction
Gap in Robot Foundation Models}

\author{
  Gihyuk Ko\\
  NC AI\\
  \texttt{gihyukko@ncsoft.com}
}

\begin{document}
\maketitle

\input{00-abstract}
\input{01-intro}
\input{02-fault}
\input{03-gap}
\input{04-agenda}
\input{05-conclusion}

\acknowledgments{This work was supported by National IT Industry Promotion
Agency (NIPA) grant funded by the Korea government(MSIT) (Development
of Physical Intelligence Action Models and an Integrated Framework, No.
RS-2026-25621677).}


\bibliographystyle{styles/corlabbrvnat}
\bibliography{bib/caus}  

\end{document}

%% file: 00-abstract.tex
\begin{abstract}
Post-incident analysis of a robot foundation model poses a \emph{singular}
causal question: which internal components caused \emph{this} failure, and to
what degree? Mechanistic interpretability (MI), however, establishes
\emph{general} causal claims by decoding the concepts a policy has learned,
yet knowing which concepts are encoded is insufficient for identifying the
actual cause of a specific incident. In this position paper, we analyze this
\emph{concept--malfunction gap} by posing fault isolation in the framework of
\emph{actual causality}. Specifically, we formalize a fault-isolation query as
a tuple of intervened internal components, counterfactual values, contingencies,
and read-outs, and lay out conditions for it to be well-specified. Through this
lens, we identify shortcomings of current MI approaches: they assume fixed
contingencies and read out only immediate effects, letting backups obscure
actual causes. We demonstrate on a toy example that these shortcomings can lead
to conflicting verdicts. Lastly, based on the well-specification conditions, we
outline four open research tasks toward closing the gap.
\end{abstract}

\keywords{Actual Causality, Fault Isolation, Robot Foundation Models,
Mechanistic Interpretability}


%% file: 01-intro.tex
\section{Introduction}
\label{sec:intro}

As robot foundation models (RFMs)---e.g., vision-language-action models
(VLAs)~\citep{zitkovich2023rt2,kim2024openvla,black2025pi0} and world action
models (WAMs)~\citep{ye2026wam}---enter spaces shared with humans, their
failures can become serious incidents. Recent legislation such as the EU AI Act
requires providers of high-risk AI systems, which can include autonomous robots,
to investigate and report such incidents and implement corrective
actions~\citep{euaiact2024}. Since retraining the entire model is costly and may
not even resolve the underlying fault, efficient compliance requires localizing
the fault for targeted repair, which comes down to one question: \emph{which
internal components actually caused the failure, and to what degree?}

The closest toolkit for analyzing the inner workings of a foundation model is
mechanistic interpretability
(MI)~\citep{zhang2026embodied,mai2026beyond,swann2026saevla}. However, MI is
primarily designed to establish \emph{general} causal claims across a
distribution~\citep{geiger2025abstraction}, whereas an individual incident calls
for a \emph{singular} one. This discrepancy creates a \emph{concept--malfunction
gap}: MI reveals which concepts a policy encodes, not which components caused
the malfunction. The closed loop of RFMs widens this gap: unlike in language
models, internal errors propagate through the environment and compound over
time.

In this position paper, we argue that the concept--malfunction gap stems
primarily from misspecified causal queries rather than deficient methods. To
re-pose these queries, we ground fault isolation in the framework of
\emph{actual causality}~\citep{halpern2005causes,halpern2016actual}, assuming
that the incident can be replayed in a deterministic simulator to compute the
counterfactuals it requires.\footnote{Unlike fault diagnosis with analytic plant
models~\citep{isermann2005fdi} or actual causality over explicit state--action
models (model checking~\citep{beer2009counterexamples},
Dec-POMDPs~\citep{triantafyllou2022decpomdp}), we seek causes among policy
internals in an unmodeled environment.} Our contributions are threefold: we
(1)~formally define a fault-isolation query $\query$
(Section~\ref{sec:setting}); (2)~diagnose how existing MI methods instantiate
$\query$ via a toy example (Section~\ref{sec:gap}); and (3)~outline four open
research tasks toward fault isolation in RFMs (Section~\ref{sec:agenda}).

%% file: 02-fault.tex
\section{Fault Isolation as Actual Causation}
\label{sec:setting}

In this section, we formalize fault isolation with actual
causality~\citep{halpern2016actual} by treating an RFM's closed-loop rollout as
a structural causal model and introducing the \emph{fault-isolation
query}~$\query$.

\textbf{Robot foundation models.} We model an RFM (VLA or WAM) as a policy
$\pi_\theta$ that maps an observation history $o_{\le t}$ (e.g., images and
proprioception) and an instruction $\ell$ to an action chunk $A_t \in
\mathbb{R}^{H \times d}$ consisting of $H$ future $d$-dimensional actions
(either discrete~\citep{zitkovich2023rt2,kim2024openvla} or
continuous~\citep{black2025pi0}), and additional future states for WAMs. We
decompose the computation of $\pi_\theta$ at step $t$ into \emph{components}
$Z_t = (Z_t^1, \dots, Z_t^n)$, the intermediate units an investigator can
intervene on and blame, e.g., attention-head outputs, MLP neurons, dictionary
latents, or a WAM's predicted-state latents.

\textbf{Incidents and fault isolation.} The exogenous context $u$ (initial
state, seed, instruction), a deterministic simulator, and $\pi_\theta$
determine the rollout $\tau$. A \emph{failure predicate}
$\varphi(\tau) \in \{0,1\}$ encodes the incident, e.g., the gripper coming
within a safety distance of a human hand, or a grasped knife being dropped.
Since each action shapes later observations, a component affects $\varphi$
\emph{directly} through its action and \emph{indirectly} through the
environment. With $V = \{Z_t^i\}_{t,i}$ collecting all components, fault
isolation asks which subset $X \subseteq V$ caused $\varphi(\tau) = 1$ in
\emph{this} rollout, not how the policy behaves on average.

\textbf{Counterfactuals and contingencies.} Actual causality builds on
\emph{but-for} dependence: \emph{had $X$ taken another value, would the failure
have been averted?} To make this precise, we treat the unrolled rollout as a
structural causal model $M$~\citep{halpern2005causes} over $V$ and $\varphi$,
with equations given by $\pi_\theta$ and the simulator. Under $M$, an
intervention $[X \leftarrow x']$ sets $X$ to $x'$ and re-runs the rollout, and
$(M,u) \models [X \leftarrow x']\,\psi$ denotes that $\psi$ holds in the
resulting \emph{counterfactual rollout}; $X{=}x$ is thus a \emph{but-for cause}
if $(M,u) \models [X \leftarrow x']\,\neg\varphi$ for some $x'$. Yet a genuine
cause may fail to be a but-for cause when a \emph{backup component} would have
produced the failure anyway (e.g., the Hydra effect~\citep{mcgrath2023hydra}).
Actual causality therefore also tests dependence under a \emph{contingency},
which holds such backups at their observed values.\footnote{In a classic
example~\citep{halpern2016actual}, Suzy and Billy both throw rocks at a bottle;
Suzy's arrives first and shatters it. Her throw is not a \emph{but-for cause},
as Billy's rock (\emph{the backup}) would have shattered the bottle anyway, but
it is an \emph{actual cause} under the contingency that \emph{Billy's rock does
not hit}, as was factually the case.}

\textbf{Definition 1 (actual cause).} Following the modified definition of
\citet{halpern2015modification}, an assignment $X{=}x$ is an \emph{actual cause}
of $\varphi$ in context $u$ if all three conditions hold:
\begin{enumerate}[label=\textbf{AC\arabic*},leftmargin=*,nosep]
  \item (factuality) $X{=}x$ and $\varphi$ both hold in the observed rollout;
  \item (counterfactual dependence) for some contingency $W \subseteq V
    \setminus X$ with observed values $w^\ast$ and some alternative value
    $x'$, $(M,u) \models [X \leftarrow x',\, W \leftarrow w^\ast]\,\neg\varphi$,
    i.e., the failure would have been averted had $X$ been $x'$ and $W$ not
    responded;
  \item (minimality) no strict subset of $X$ satisfies AC1 and AC2.
\end{enumerate}
Such $W$ is called a \emph{witness}~\citep{halpern2016actual}, and but-for
causation is the case $W = \emptyset$. Holding $W$ at $w^\ast$ amounts to
activation patching~\citep{zhang2024patching}, where factual activations are
used during the counterfactual rollout.

\textbf{Definition 2 (degree of responsibility).} Following
\citet{chockler2004responsibility}, as adapted to the modified
definition~\citep[Ch.~6]{halpern2016actual}, the \emph{degree of
responsibility} of an assignment $Z{=}z$ with $Z \in V$ for $\varphi$ is $1/k$,
where $k$ is the minimum of $|X| + |W|$ over all actual causes $X{=}x$ that
include $Z{=}z$ and their witnesses $W$; it is $0$ if no such cause exists.

Definition~1 is binary: it cannot distinguish a cause with one backup from a
cause with ten, nor rank candidate causes for repair. Definition~2 grades them,
as $k$ counts the components that must be changed ($X$) or held fixed ($W$) for
$\varphi$ to depend on $Z$. A single-variable but-for cause thus has
responsibility $1$, whereas a cause with one or ten backups has $1/2$ or
$1/11$. A component changed or held at several steps counts as a single
variable, both in $k$ and in AC3.

\textbf{Definition 3 (fault-isolation query).} A \emph{fault-isolation query}
$\query = \langle X, x', W, Y \rangle$ consists of intervened components $X
\subseteq V$, a counterfactual value $x'$, a contingency $W \subseteq V
\setminus X$, and a read-out $Y$; it tests whether $Y$ changes under $[X
\leftarrow x', W \leftarrow w^\ast]$. We call $\query$ \emph{well-specified} for
an RFM incident if all four conditions hold, and \emph{misspecified}
otherwise:
\begin{enumerate}[label=\textbf{WS\arabic*},leftmargin=*,nosep]
  \item (temporal scope) $X$ may include components at any step;
  \item (realizability) $x'$ can be induced by some observation history;
  \item (searched contingency) $W$ is searched per incident, not fixed in
    advance;
  \item (rollout outcome) $Y$ is $\varphi$ evaluated on the counterfactual
    rollout.
\end{enumerate}

A misspecified $\query$ can reach a wrong verdict: a single-step $X$ misses
faults that accumulate over steps (WS1); an unrealizable $x'$ may avert
$\varphi$ merely by driving the policy off-manifold (WS2; cf.\
normality~\citep{halpern2015graded}); a fixed $W$ may hide causes that have
backups, or misgrade them (WS3); and a per-step read-out can miss effects
mediated by the environment (WS4).

\textbf{Handling backups in RFMs.} (i)~\emph{Loop backups and realizability:} in
RFMs, the policy itself can undo a repair at later steps (e.g., a gripper
stopped short of its goal simply moves toward it again), so a witness must also
freeze such later reactions. A cause
revealed this way is a \emph{persistent} fault: repairing it alone will not
prevent the failure. Such witnesses, however, may force combinations the policy
would never produce (e.g., a slowed robot whose drive stays idle), so their
verdicts weaken as the counterfactual drifts from the factual rollout.
(ii)~\emph{Granularity and responsibility:} responsibility (Definition~2) counts
how many components must be frozen, which is meaningful only if each backup is
one component; sparse autoencoder latents~\citep{huben2024sae} approximate this
better than individual neurons.

%% file: 03-gap.tex
\section{The Concept--Malfunction Gap in Current MI Methods}
\label{sec:gap}

\input{figures/tab-gap}

In this section, we view existing MI methods as instantiations of $\query$,
check whether they are well-specified, and illustrate how their
misspecifications may lead to conflicting verdicts via an example.

\textbf{Existing MI methods as instantiations of $\query$.} We examine six
MI method families and summarize in Table~\ref{tab:gap} how each instantiates
$\query$. None yields a well-specified query. Violating only WS1 or WS3 would
merely miss causes, but every intervening family also violates WS2, so even
its reported causes may be spurious. Specifically, they violate temporal
scope (WS1) by intervening within a single step or only on the input (except
steering); realizability (WS2) by using counterfactual values that the incident
may never produce, such as masked images, other
inputs~\citep{zhang2024patching}, or steered states; searched contingency (WS3)
by fixing $W$ in advance or averaging it over inputs, thus ignoring the
incident's backups; and rollout outcome (WS4) by reading out a per-step
surrogate such as the action $A_t$ (except visual attribution and steering),
which misses what the loop does next.\footnote{Some of these gaps are among the
open problems in MI identified by \citet{sharkey2025open}: single-step analysis
(WS1), backup behavior (WS3), and direct-effect-only read-outs (WS4).}

\textbf{A toy example.} In our scenario, a policy drives a robot toward a human
$2$\,m away at $v_{\mathrm{ref}} = 1$\,m/s (Figure~\ref{fig:query}a; simulated
at $10$\,Hz for $5$\,s), and the incident $\varphi = 1$ occurs if the robot
comes within $0.3$\,m of the human. The policy sets the acceleration $a_t =
-2\,h_t(1-s_t) + g_t$ through three components: a hazard latent $h_t = 1$ that
fires when a person is in view and triggers braking; a suppressor $s_t \in
[0,1]$ that blocks this braking while a task is active; and a target-reaching
drive $g_t = 4 \max(0, v_{\mathrm{ref}} - v_t)$ that accelerates whenever the
robot falls below its reference speed. By design, the suppressor is always on
($s_t = 1$), so the robot never brakes or triggers the drive ($g_t = 0$),
breaching at $1.7$\,s. The suppressor $s$ is thus an actual cause: a sound
method should blame $s$, not the hazard latent $h$ (active, but blocked by $s$)
or the drive $g$ (idle throughout). Yet $g$ acts as a backup; even if $s$ were
repaired, $g$ would keep driving the robot toward the human until it breaches,
so a fix must address both $s$ and $g$.

\input{figures/fig-query}

\textbf{Conflicting verdicts.} We now demonstrate that applying the MI methods
in Table~\ref{tab:gap} yields conflicting verdicts. First, \emph{probing} finds
the hazard encoded in $h$, yet cannot explain why the policy failed to brake
even though it ``saw'' the human. \emph{Visual attribution} sees no change when
the human is masked, as $s$ already blocks braking regardless of $h$, and in any
case blames pixels rather than internal components. Next, single-pass
\emph{activation patching} of $s_0 \leftarrow 0$ (Figure~\ref{fig:query}b)
initially yields full braking ($|\Delta a_0| = 2$\,m/s$^2$), but a
counterfactual rollout reveals that the breach still occurs, only delayed by
$0.1$\,s (WS1, WS4). Patching $s$ at all steps also fails to isolate $s$ as a
cause, as the backup $g$ settles the robot at $0.5$\,m/s, still resulting in a
breach at $3.3$\,s (WS3); \emph{causal scrubbing} and \emph{causal abstraction},
which also fix $W$, inherit this blind spot. Lastly, \emph{steering} $s
\leftarrow 1-\alpha$ finds no realizable $\alpha \le 1$ that averts $\varphi$,
since $g$ always restores the robot's speed (WS3). Only the \emph{well-specified
query} on $s$, freezing $g$ (Figure~\ref{fig:query}c), recovers the ground
truth: $s{=}1$ is an actual cause with responsibility $1/2$ ($X = \{s\}$, $W =
\{g\}$), albeit via an off-manifold witness (Section~\ref{sec:setting}).

%% file: figures/tab-gap.tex
\begin{table}[t]
\centering
\caption{The requirements of a well-specified query (top), existing MI methods
as instantiations of $\query$ (middle), and open research tasks toward meeting
them (bottom). \cmark/\xmark: met/not met.}
\label{tab:gap}
\scriptsize
\setlength{\tabcolsep}{4.5pt}
\setlength{\aboverulesep}{0.25ex}\setlength{\belowrulesep}{0.3ex}
\begin{tabular}{@{}lllll@{}}
\toprule
\textbf{Condition} $\rightarrow$ & \textbf{WS1} & \textbf{WS2} & \textbf{WS3} & \textbf{WS4} \\
\textbf{Method} $\downarrow$ & \emph{scope of $X$} & \emph{realizability of $x'$} & \emph{flexibility of $W$} & \emph{horizon of $Y$} \\
\midrule\midrule
\textbf{Well-specified} & \textbf{across steps} & \textbf{realizable} & \textbf{searched per incident} & \textbf{failure $\varphi$, full rollout} \\
\midrule
Probing / SAEs~\citep{mai2026beyond,swann2026saevla,huben2024sae} & \xmark~no intervention & \xmark~--- & \xmark~--- & \xmark~latent value, one step \\
Visual attribution~\citep{zhang2026embodied} & \xmark~input image only & \xmark~masking & \xmark~$\emptyset$ & \cmark~action or rollout \\
Activation / path patching~\citep{wang2023ioi,zhang2024patching} & \xmark~one step & \xmark~other input & \xmark~$\emptyset$ or paths & \xmark~logits, one step \\
Causal scrubbing~\citep{chan2022scrubbing} & \xmark~one step & \xmark~resampling & \xmark~set by hypothesis & \xmark~loss, one step \\
Activation steering~\citep{turner2023steering,rimsky2024caa,haon2025steering} & \cmark~all steps & \xmark~steering & \xmark~$\emptyset$ & \cmark~rollout behavior \\
Causal abstraction~\citep{geiger2025abstraction} & \xmark~one step & \xmark~source input & \xmark~averaged over inputs & \xmark~high-level var., one step \\
\midrule\midrule
\textbf{Open research tasks} & \textbf{T1}: finding faults & \textbf{T2}: certifying $x'$ & \textbf{T3}: scaling contingency search & \textbf{T1}: finding faults \\
                  & \multicolumn{4}{c}{\textbf{T4}: validating fault isolation (all conditions jointly)} \\
\bottomrule
\end{tabular}
\end{table}

%% file: figures/fig-query.tex
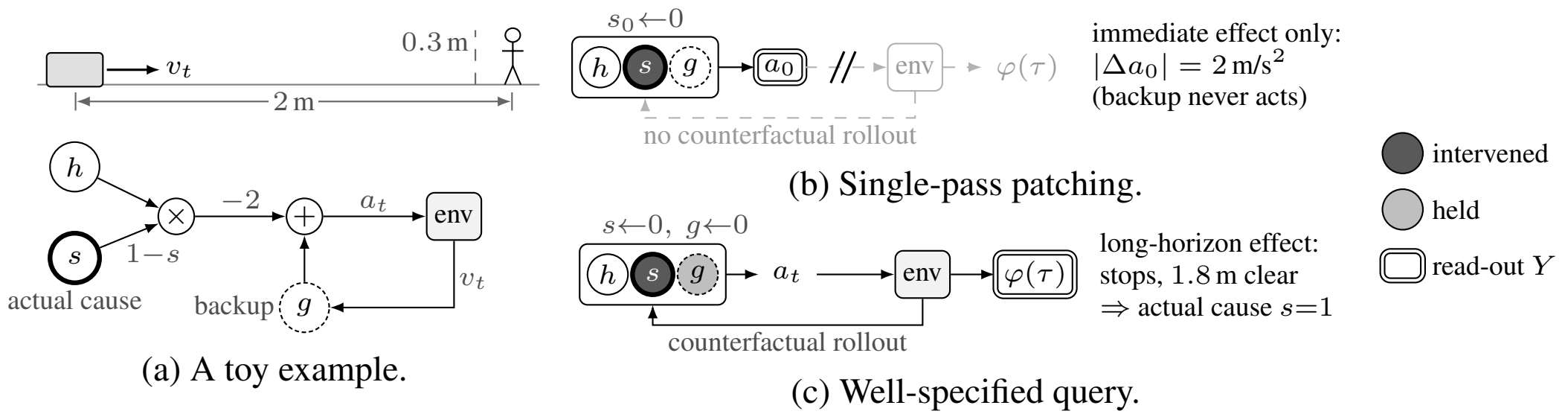
\begin{figure}[t]
\centering
\tikzset{
  lat/.style={circle,draw,minimum size=4.4mm,inner sep=0pt},
  op/.style={circle,draw,minimum size=3.2mm,inner sep=0pt},
  cmp/.style={circle,draw,minimum size=3.6mm,inner sep=0pt},
  xnode/.style={fill=black!65,text=white},
  wnode/.style={fill=black!25},
  causenode/.style={very thick},
  backupnode/.style={dash pattern=on 1.1pt off 0.7pt},
  envb/.style={draw,rounded corners=2pt,fill=black!5,minimum height=4.2mm,inner sep=2pt},
  rdout/.style={draw,double,double distance=0.7pt,rounded corners=2pt,inner sep=2pt},
  dimb/.style={draw=black!30,text=black!40,fill=none},
  ardim/.style={->,draw=black!30,dashed},
  lbl/.style={font=\scriptsize,text=black!70,inner sep=1pt}}
\newcommand{\polblock}[2]{%
  \draw[rounded corners=2pt] (0,#1-0.27) rectangle (1.3,#1+0.27);
  \node[cmp] at (0.25,#1) {$h$};
  \node[cmp,causenode,xnode] at (0.65,#1) {$s$};
  \node[cmp,backupnode,#2] at (1.05,#1) {$g$};}
\newcommand{\subcap}[1]{\par\vspace{1pt}{\footnotesize #1}}
\begin{minipage}[c]{0.35\textwidth}
\centering
\begin{tikzpicture}[font=\scriptsize,>={Latex[length=1.2mm]},x=0.95cm,y=0.88cm]
\draw[black!50] (-0.15,1.95) -- (4.55,1.95);
\node[draw,fill=black!12,rounded corners=1pt,minimum width=5mm,minimum height=3mm,inner sep=0pt] at (0.2,2.1) {};
\draw[->,semithick] (0.5,2.1) -- (1.0,2.1) node[right,inner sep=1pt] {$v_t$};
\begin{scope}[shift={(4.3,1.95)}]
  \draw (0,0.5) circle (0.07);
  \draw (0,0.43) -- (0,0.19) (0,0.19) -- (-0.075,0) (0,0.19) -- (0.075,0) (-0.1,0.34) -- (0.1,0.34);
\end{scope}
\draw[dashed,black!60] (3.95,1.95) -- (3.95,2.5);
\node[lbl,anchor=east] at (3.93,2.38) {$0.3$\,m};
\draw[|<->|,black!60] (0.2,1.78) -- node[lbl,fill=white] {$2$\,m} (4.3,1.78);
\node[lat] (h) at (0.2,1.12) {$h$};
\node[lat,causenode] (s) at (0.2,0.2) {$s$};
\node[lbl,anchor=north] at (s.south) {actual cause};
\node[op] (m) at (1.15,0.62) {$\times$};
\draw[->] (h) -- (m);
\draw[->] (s) -- node[lbl,pos=0.35,below right=0pt] {$1{-}s$} (m);
\node[op] (sum) at (2.35,0.62) {$+$};
\draw[->] (m) -- node[lbl,above] {$-2$} (sum);
\node[lat,backupnode] (g) at (2.35,-0.3) {$g$};
\node[lbl,anchor=east] at (g.west) {backup};
\draw[->] (g) -- (sum);
\node[envb] (env) at (3.75,0.62) {env};
\draw[->] (sum) -- node[lbl,above] {$a_t$} (env);
\draw[->] (env.south) |- node[lbl,pos=0.3,right] {$v_t$} (g.east);
\end{tikzpicture}
\subcap{(a) A toy example.}
\end{minipage}\hfill
\begin{minipage}[c]{0.5\textwidth}
\centering
\begin{tikzpicture}[font=\scriptsize,>={Latex[length=1.2mm]}]
\polblock{0}{}
\node[lbl,anchor=south] at (0.65,0.28) {$s_0{\leftarrow}0$};
\node[rdout] (aA) at (1.85,0) {$a_0$};
\draw[->] (1.3,0) -- (aA);
\node[envb,dimb] (aE) at (3.05,0) {env};
\draw[ardim] (aA) -- (aE);
\draw[semithick] (2.3,-0.14) -- (2.42,0.14) (2.4,-0.14) -- (2.52,0.14);
\node[text=black!40] (aF) at (4.05,0) {$\varphi(\tau)$};
\draw[ardim] (aE) -- (aF);
\draw[ardim] (aE.south) |- (0.65,-0.45) -- (0.65,-0.27);
\node[lbl,anchor=north,text=black!40] at (1.85,-0.47) {no counterfactual rollout};
\node[anchor=west,align=left] at (4.5,0)
  {immediate effect only:\\$|\Delta a_0| = 2$\,m/s$^2$\\(backup never acts)};
\end{tikzpicture}
\subcap{(b) Single-pass patching.}
\par\vspace{1pt}
\begin{tikzpicture}[font=\scriptsize,>={Latex[length=1.2mm]}]
\polblock{0}{wnode}
\node[lbl,anchor=south] at (0.85,0.28) {$s{\leftarrow}0,\ g{\leftarrow}0$};
\node (bA) at (1.85,0) {$a_t$};
\draw[->] (1.3,0) -- (bA);
\node[envb] (bE) at (3.05,0) {env};
\draw[->] (bA) -- (bE);
\node[rdout] (bF) at (4.05,0) {$\varphi(\tau)$};
\draw[->] (bE) -- (bF);
\draw[->] (bE.south) |- (0.65,-0.45) -- (0.65,-0.27);
\node[lbl,anchor=north] at (1.85,-0.47) {counterfactual rollout};
\node[anchor=west,align=left] at (4.5,0)
  {long-horizon effect:\\stops, $1.8$\,m clear\\$\Rightarrow$ actual cause $s{=}1$};
\end{tikzpicture}
\subcap{(c) Well-specified query.}
\end{minipage}\hfill
\begin{minipage}[c]{0.12\textwidth}
\begin{tikzpicture}[font=\scriptsize]
\node[cmp,xnode] (l1) at (0,0.5) {};
\node[anchor=west,inner sep=2pt] at (l1.east) {intervened};
\node[cmp,wnode] (l2) at (0,0) {};
\node[anchor=west,inner sep=2pt] at (l2.east) {held};
\node[rdout,minimum width=3.6mm,minimum height=2.6mm] (l3) at (0,-0.5) {};
\node[anchor=west,inner sep=2pt] at (l3.east) {read-out $Y$};
\end{tikzpicture}
\end{minipage}
\caption{(a)~A toy example: the suppressor $s$ (actual cause) blocks braking on
the hazard latent $h$, and the drive $g$ (its backup) re-accelerates the robot
through the loop. (b)~Single-pass patching reads out only the immediate effect of the
intervention; (c)~the well-specified query, with contingency $g \leftarrow 0$,
correctly identifies $s{=}1$ as an actual cause.}
\label{fig:query}
\end{figure}

%% file: 04-agenda.tex
\section{Open Research Tasks}
\label{sec:agenda}

In this section, we identify four open research tasks, each arising from
meeting a well-specification condition of $\query$ (Definition~3) at the scale
of real RFMs (Table~\ref{tab:gap}, bottom).

\textbf{Task 1: Finding candidate faults (WS1, WS4).} An investigator must
decide which components to test as faults ($X$), as testing every component
is intractable. As modern RFMs generate continuous action
chunks~\citep{black2025pi0}, this may require an \emph{action lens}, a
counterpart of the logit lens~\citep{nostalgebraist2020logit} that projects
components onto problematic actions; its candidates must then be tested in
counterfactual rollouts.

\textbf{Task 2: Certifying realizable counterfactuals (WS2).} In an RFM, a
patched activation may not be induced by any observation history. Alternatively,
activation steering along a direction $\delta$ separating safe from unsafe
behavior ($x' = x + \alpha\delta$) makes the counterfactual more likely to stay
within the policy's representations, but offers no guarantee that any
observation history actually induces it.

\textbf{Task 3: Scaling contingency search (WS3).} Given a candidate fault, its
contingency ($W$) must be found. In an RFM, this means searching among
millions of latents, including dormant ones. As deciding actual causation is
$D^P$-complete even for explicit models~\citep{halpern2015modification},
candidate backups must be shortlisted, e.g., by
attribution~\citep{syed2024attribution}.

\textbf{Task 4: Validating fault isolation.} Judging whether a fault was
correctly isolated requires knowing which components caused the incident.
Naturally trained RFMs rarely offer this, as no fault is built in by design.
Planting faults with known causes and backups, e.g., via interchange
intervention training~\citep{geiger2022iit}, could provide ground truth, but planted faults must be as realistic as natural ones.

%% file: 05-conclusion.tex
\section{Conclusion}
\label{sec:conclusion}

In this paper, we addressed the concept--malfunction gap in RFMs: the
discrepancy between the concepts that interpretability research aims to
uncover and the actual causes of a specific malfunction. Formalizing fault
isolation with actual causality revealed that the gap stems primarily from how
existing MI methods specify the causal query rather than from their tools.
Specifically, they assume fixed contingencies and read out only immediate
effects, which lets backup components affect the counterfactual rollout and
obscure actual causes. Answering well-specified queries at the scale of real
RFMs still requires new tools, for which we outlined four open research tasks.

\textbf{Limitations.} As this is a position paper, we provide no experiments on
real RFMs: our argument rests on a toy example and assumes a faithful simulator,
leaving empirical validation to future work.